\documentclass[accepted]{article}

\usepackage{microtype}
\usepackage{graphicx}
\usepackage{subfigure}
\usepackage{makecell}
\usepackage{booktabs} % for professional tables
\usepackage{pgfplots}
    \usepgfplotslibrary{groupplots}
    \usetikzlibrary{matrix}

\usepackage{hyperref}

\usepackage{icml2020}

\usepackage{amsmath, amsfonts, amssymb}
\usepackage{bm}
\usepackage{tikz}
\usetikzlibrary{calc}
\usepackage{pgfplotstable}
\usepackage{pgfplots}
\usepackage{array,multirow,graphicx}
\usepackage{tabularx,colortbl}
\usepackage[colorinlistoftodos,prependcaption]{todonotes}
\usepackage{enumitem}
\usepackage{standalone}
\pgfplotsset{compat=1.13}

\def \y{y}
\def \x{x}
\def \w{w}
\def \g{g}

\def \R{\mathbb{R}}			% real numbers
\def \N {\mathbb{N}}			% nat. numbers

\DeclareMathAlphabet{\pazocal}{OMS}{zplm}{m}{n}

\newcommand{\algorithmicconstants}{\textbf{Constants:}}
\newcommand\CONSTANTS{\item[\algorithmicconstants]}
\newcommand{\algorithmicreturn}{\textbf{return: }}
\newcommand\RETURN{\item[\algorithmicreturn]}
\newcommand{\algorithmiccommentMine}[1]{\bgroup\hfill$\triangleright$~\emph{#1}\egroup}

\usepackage{amsthm}

\usetikzlibrary{calc}
\usepackage{stmaryrd}
\usepackage{upgreek}
\usetikzlibrary{shapes.geometric}
\usetikzlibrary{decorations.pathreplacing}

\definecolor{myblack}{RGB}{53, 53, 53}
\definecolor{myblue}{RGB}{40, 75, 200}
\definecolor{myred}{RGB}{192, 50, 33}
\definecolor{myyellow}{RGB}{255, 166, 48}
\definecolor{mywhite}{RGB}{240, 237, 238}
\definecolor{mygreen}{RGB}{0, 102, 0}
\definecolor{mypurple}{RGB}{150, 0, 180}

\definecolor{green1}{RGB}{9, 82, 86}
\definecolor{green2}{RGB}{8, 127, 140}
\definecolor{green3}{RGB}{6, 167, 125}
\definecolor{green4}{RGB}{79, 109, 122}
\definecolor{green5}{RGB}{192, 214, 223}
\definecolor{violet}{RGB}{26,69,131}

\newcommand\oldtext[1]{}

\icmltitlerunning{
A Multilevel Approach to Training
}

\begin{document}

\twocolumn[
% Better ideas ???
\icmltitle{
A Multilevel Approach to Training
}



\icmlsetsymbol{equal}{*}

\begin{icmlauthorlist}
\icmlauthor{Vanessa Braglia}{equal,usi}
\icmlauthor{Alena Kopani\v{c}\'akov\'a}{equal,usi}
\icmlauthor{Rolf Krause}{usi}
\end{icmlauthorlist}

\icmlaffiliation{usi}{Institute of Computational Science, Universit\`a della Svizzera italiana}
\icmlcorrespondingauthor{Alena Ktopani\v{c}\'akov\'a}{alena.kopanicakova@usi.ch}

% You may provide any keywords that yout
\icmlkeywords{multilevel minimization, machine learning, subsampling, stochastic optimization}
\vskip 0.3in
]



\printAffiliationsAndNotice{\icmlEqualContribution} % otherwise use the standard text.

\begin{abstract}
We propose a novel training method based on nonlinear multilevel minimization techniques, commonly used for solving discretized large scale partial differential equations.   
Our multilevel training method constructs a multilevel hierarchy by reducing the number of samples. 
The training of the original model is then enhanced by internally training surrogate models constructed with fewer samples. 
We construct the surrogate models using first-order consistency approach. 
This gives rise to surrogate models, whose gradients are stochastic estimators of the full gradient, but with reduced variance compared to standard stochastic gradient estimators. 
We illustrate the convergence behavior of the proposed multilevel method to machine learning applications based on logistic regression. 
A comparison with subsampled Newton's and variance reduction methods demonstrate the efficiency of our multilevel method. 
\end{abstract}

%%%%%%%%%%%%%%%%%%%%%%%%%%%%%%%%%%%%%%%%%
%%%%%%%%%%%%%%%%%%%%%%%%%%%%%%%%%%%%%%%%%
%%%%%%%%%%%%%%%%%%%%%%%%%%%%%%%%%%%%%%%%%
\section{Introduction}
\label{sec:introduction}
We consider the following minimization problem
\begin{align}
\min_{\w \in \R^d} \pazocal{F}(\w) = \frac{1}{n} \sum_{j=1}^{n} f_{j} (\w),
\label{eq:min_problem}
\end{align}
where each $f_j: \R^d \rightarrow \R$ is smooth and convex.
Problems of this type arise frequently in supervised learning applications, such as logistic or least squares regression \cite{goodfellow2016deep}. 
Minimizing \eqref{eq:min_problem} using standard deterministic methods, such as gradient descent (GD) or Newton's method, is often prohibitive in practice, especially for large datasets \cite{bottou2018optimization}.
A popular alternative used by machine learning practitioners is stochastic gradient descent (SGD) \cite{robbins1951stochastic}, which uses an unbiased gradient estimator. 
The main drawback of the SGD method is its sensitivity to the variance of gradient estimates, which prevents SGD from converging to a minimizer when fixed stepsizes are used.
To overcome this difficulty, a diminishing sequence of stepsizes can be used, which leads to slower convergence.

To address these limitations of SGD, variance reduction (VR) methods can be employed.
The main idea behind VR methods is to combine deterministic and stochastic aspects in order to decrease the variance of the stochastic gradient estimator.
Representative algorithms of this class are for example
SAG \cite{schmidt2017minimizing}, 
SAGA \cite{defazio2014saga}, 
S2GD \cite{konevcny2017semi},
SVRG \cite{johnson2013accelerating},
MISO  \cite{mairal2013optimization}, 
SARAH \cite{nguyen2017sarah}.
Although these VR methods show strong theoretical and practical results for convex optimization problems, their convergence rate often deteriorates, when the underlying problem is ill-conditioned.

Multilevel methods are well known, in numerical analysis, to address issues related to ill-conditioning, as they provide accurate approximation to inverse of the Hessian.
Hence, they can be interpreted as a second order approach.
The introduction into multilevel methods can be found for example in \cite{Briggs2000multigrid, hackbusch2013multi}. 
An extension to nonlinear problems was originally proposed in \cite{brandt1977multi}, which led to many developments concerning the convex minimization problems, such as \cite{Nash2000multigrid, kornhuber_adaptive_2001, tai2002global, chenConvergenceAnalysisFast2019}, 
as well as non-convex minimization problems, e.g.  \cite{Gratton2008recursive, Gross2009, kopanicakova2019b}.

Motivated by the effectiveness of variance reduction and multilevel methods, we propose a multilevel variance reduction (MLVR) method, which combines both aspects. 
Our MLVR method can be seen as a variant of MG/OPT \cite{Nash2000multigrid}, developed for minimizing \eqref{eq:min_problem}. 
By design, our MLVR method constructs a multilevel hierarchy by reducing the number of samples.
The convergence of the original problem is then enhanced by internally minimizing the surrogate models based on sub-sampled data.
Since the surrogate models are constructed by combining deterministic and stochastic information, their gradients have lower variance than gradients arising from purely stochastic settings (SGD method). 
Indeed, we demonstrate in Section \ref{sec:connection} that our MLVR method can be configured in such a way, that it degenerates to already known VR methods in the machine learning community.

The presented MLVR method employs a multilevel hierarchy created by reducing the number of samples, while the number of parameters is kept fixed.
This is very convenient, as it makes MLVR applicable to a wide range of machine learning models.
In contrast, other recently developed multilevel methods in the machine learning community are not as flexible, as they assume a particular structure of the underlying problem, see for example \cite{hovhannisyan2016magma, chang2017multi,  gaedke2020multilevel}.

%%%%%%%%%%%%%%%%%%%%%%%%%%%%%%%%%%%%%%%%%
%%%%%%%%%%%%%%%%%%%%%%%%%%%%%%%%%%%%%%%%%
%%%%%%%%%%%%%%%%%%%%%%%%%%%%%%%%%%%%%%%%%
\section{Multilevel Training}
\label{sec:multigrid}
In this section, we propose a multilevel variance reduction method (MLVR) for minimizing problems of type \eqref{eq:min_problem} arising in supervised learning applications.
We assume that the dataset  $\pazocal{D} = \{ (\x_j, \y_j) \}_{j=1}^n$ of $n$ samples is given and each sample is represented by feature vector $\x_j \in \R^d$ and respective label $\y_j \in \R$.
The proposed MLVR method can be seen as a variant of the MG/OPT method \cite{Nash2000multigrid}, specifically tailored for the problem at the hand, where $n$ is usually large.
The main idea behind nonlinear multilevel methods is to create the hierarchy of $L$ levels. 
Each level $l \in \{1, \dots, L \}$ is then associated with minimization of some auxiliary level dependent objective function $\pazocal{H}^l: \R^{d^l} \rightarrow \R$, where $d^l \leq d^{l+1}$. 
On the finest level, where $l=L$, we identify $\pazocal{H}^{L}$ with our target objective function, thus $\pazocal{H}^{L} (\w)= \pazocal{F}(\w)$, for all $\w \in \R^d$.

\subsection{Multilevel Hierarchy (Coarsening in Samples)}
The level dependent objective functions are constructed in such a way, that they are computationally cheaper to minimize than $\pazocal{F}$. 
We construct low-cost approximations of $\pazocal{F}$ by reducing the number of samples.
To this aim, we create hierarchy of datasets $\{ \pazocal{D}^l \}_{l=1}^{l=L}$, such that 
\begin{align}
\pazocal{D}^1 \subseteq \pazocal{D}^2  \subseteq  \dots \subseteq   \pazocal{D}^{L-1} \subseteq \pazocal{D}^{L} := \pazocal{D}.
\label{eq:data_sets}
\end{align}
Thus, the finest level, $l=L$, is associated with the full dataset $\pazocal{D}$, while the cardinality of the dataset decreases on lower levels,  i.e. $ |\pazocal{D}^{l-1}| \leq |\pazocal{D}^l |$.
There are several possibilities how to obtain hierarchy of datasets $\{ \pazocal{D}^l \}_{l=1}^{l=L}$, such that \eqref{eq:data_sets} is satisfied. 
Here, we construct $\pazocal{D}^l$ by randomly choosing samples from 
$\pazocal{D}^{l+1}$ 
in uniform manner.

\paragraph{Transfer Operators}
The multilevel methods necessitate transfer of data between subsequent levels of the multilevel hierarchy. 
The MLVR method proposed in this work is based a the coarsening in the samples, while the parameter space is fixed.
This is very convenient, as the transfer operators, known in multilevel literature as prolongation and restriction operators, become identity - even in their algebraic forms. 
As a consequence, the practical implementation of MLVR method is simplified, compared to traditional nonlinear multilevel minimization methods.

\subsection{The Training (MLVR Algorithm)}
We present MLVR algorithm in the form of a V-cycle, which consists of a downward and an upward phase. 
The algorithm begins on the finest level, $l=L$, with some initial parameters $\w_0^L$.
During the downward phase, we pass through all levels of the multilevel hierarchy until the coarsest level is reached.
On every level, we approximately minimize level-dependent objective function $\pazocal{H}^l$ by performing $\mu_1^l$ steps of some level dependent optimizer.
The updated parameters, $\w_{\mu_1^l}^l$, are then used as an initial guess for subsequent coarser level, i.e $\w_0^{l-1} =\w_{\mu_1^l}^l$.
Once the coarsest level is reached, the MLVR performs $\mu^1$ level-1-optimizer steps. 
Updated parameters, $\w_{\mu^2}^1$, are then transferred to the finer level, i.e. $\w_{\mu_1+1}^2 = \w_{\mu^2}^1$, where they are again updated by executing $\mu^l_2$ steps of the level-dependent optimizer.
This process is repeated until the finest level is reached, see  Algorithm \ref{alg:mgopt}. 

\begin{algorithm}
	\caption{\footnotesize V-cycle of MLVR($l, \w^{l}_{0}, \delta {g}^{l} $) }
	\footnotesize
	\label{alg:mgopt}
	\begin{algorithmic}
	\CONSTANTS $ \mu_1^l, \mu_2^l, \mu^1 \in \N$
		\STATE \textit{1. Downward phase}
		\STATE \quad Construct dataset $\pazocal{D}^l$ and objective function $\pazocal{H}^{l}$
		\STATE \quad  $[\w^{l}_{\mu_1^l}]$ =  LevelOptimizer($\mathcal{H}^{l}$, $\w^{l}_{0}$, $\mu_1^l$)
		\STATE \quad $\w^{l-1}_{0} \mapsfrom  \w^{l}_{\mu_1^l}$ 
		\STATE \textit{2. Recursion or call to optimizer on the coarsest level}
		 \STATE \quad \textbf{if} $l = 2$ \textbf{then}
		\STATE \quad \quad Construct dataset $\pazocal{D}^1$ and objective function $\pazocal{H}^{1}$
		 \STATE \quad \quad $[\w^{l-1}_{\mu^l}]$ =  LevelOptimizer($\pazocal{H}^{1}$, $\w^{l-1}_{0}$, $\mu^{1}$)
		 \STATE \quad \textbf{else}
		 \STATE  \quad \quad $[\w^{l-1}_{\mu^l}]$ = MLVR($l-1, \w^{l-1}_{0}, \delta {g}^{l-1} $) 
		 \STATE \quad \textbf{end if}
		\STATE \textit{3. Upward phase}		 
		\STATE \quad $\w^{l}_{ \mu^l_1 +1} \mapsfrom  \w^{l-1}_{\mu^l}$
		\STATE \quad $[\w^{l}_{\mu^l}]$ =  LevelOptimizer($\pazocal{H}^{l}$, $\w^{l}_{\mu^l +1}$, $\mu_2^l$)
		\RETURN $\w^{l}_{\mu^l}$
	\end{algorithmic}
\end{algorithm}

\begin{algorithm}
	\caption{	\footnotesize LevelOptimizer($\pazocal{H}^l$, $\w^{l}_{0}$, max\_it)}
	\label{alg:ml_alg_gd}
	\footnotesize	
	\begin{algorithmic}
		\CONSTANTS  $\alpha\in \R^+$	
		\FOR{ $i = 1, \dots, \text{max\_it}$}
		\STATE // Gradient descent step
		\STATE $\w^{l}_{i} = \w^{l}_{i-1} - \alpha \nabla \pazocal{H}^l(\w^{l}_{i-1})$
		\STATE
		\STATE // Newton step
		\STATE // $\w^{l}_{i} = \w^{l}_{i-1} - \alpha (\nabla^2 \pazocal{H}^l(\w^{l}_{i-1}) )^{-1}  \nabla \pazocal{H}^l(\w^{l}_{i-1})$
		\STATE		
		\STATE // Adam, SGD, AdaGrad, \dots, step
		\ENDFOR
		\RETURN $\w^{l}_{ \text{max\_it}}$
	\end{algorithmic}
\end{algorithm}

\subsection{Level Dependent Objective Functions}
At each level of the multilevel hierarchy, the MLVR method approximately minimizes some level dependent objective function $\pazocal{H}^l:\R^d \rightarrow \R$.
The choice of $\pazocal{H}^l$ plays a crucial role in practice, as the minimization of $\pazocal{H}^l$ should produce good search direction with respect to the fine level.
Several models were developed in the literature, see for instance \cite{alexandrov2001overview, yavneh2006multilevel, kopanicakova_2019c}. 
Here, we follow standard first-order consistency approach \cite{Nash2000multigrid, brandt1977multi}, and define $\pazocal{H}^l$ in additive manner as 
\begin{align}
\pazocal{H}^{l}(\w^{l}) := \pazocal{F}^{l}(\w^{l}) + 
\langle \delta \g^l, \w^l - \w^l_0 \rangle,
\label{eq:coarse_objective}
\end{align}
where  $\pazocal{F}^{l}: \R^d \rightarrow \R$ denotes a sub-sampled surrogate of the original objective function $\pazocal{F}$, as 
\begin{align}
\pazocal{F}^l := \frac{1}{| \pazocal{D}^l |} \sum_{j \in \pazocal{D}^l} f_j(\w).
\end{align}
The term $\delta g$ from \eqref{eq:coarse_objective}, defined as 
\begin{align}
\delta \g^l := 
\begin{cases}
\nabla \pazocal{H}^{l+1}(\w^{l+1}_{\mu_1})  - \nabla \pazocal{F}^{l}(\w^{l}_{0}), &  \text{if}  \ \ \ l < L,\\
0,  & \text{if}  \ \ \ l = L,
\end{cases}
\end{align}
ensures the first-order consistency between the coarse and fine-level objective functions at $\w^{l+1}_{\mu_1^{l+1}}$ and $\w^{l}_0$, i.e.
\begin{align}
\nabla \pazocal{H}^l(\w^l_0) = \nabla \pazocal{H}^{l+1}(\w_{\mu_1^{l+1}}^{l+1}).
\label{eq:cc_g}
\end{align}
In this way, the model $\pazocal{H}^l$ behaves as a first-order Taylor series approximation to $\pazocal{H}^{l+1}$ at points where \eqref{eq:cc_g} is satisfied. 
Hence, the local behavior of $\pazocal{H}^l$ and $\pazocal{H}^{l+1}$ is same in neighborhood of $\w^{l+1}_{\mu_1^{l+1}}$ and $\w^l_0$, respectively. 
This provides many useful properties, which we briefly discuss below.

\paragraph{Descent Directions}
By definition, the coarse level objective function $\pazocal{H}^{l}$ does not capture the underlying problem with the same accuracy as its higher-level counterpart $\pazocal{H}^{l+1}$.
However, $\pazocal{H}^{l}$ has satisfactory properties for finding search directions, which improve a fine level model.
To demonstrate this property,  let us consider some coarse level search direction $p^{l}$.
We assume that $p^{l}$ is a descent direction for $\pazocal{H}^{l}$ at $\w^{l}_{0}$, thus that
$\langle \nabla \pazocal{H}^{l}( \w^{l}_{0}), p^{l} \rangle < 0$.
Using first-order consistency relation \eqref{eq:cc_g} and fact that $\w^{l}_0 = \w^{l+1}_{\mu_1^{l+1}}$ and $p^{l+1} = p^l$, we  can also show that
$\langle \nabla \pazocal{H}^{l+1}( \w^{l+1}_{\mu_1^{l+1}}), p^{l} \rangle < 0$. 
Thus, the that search direction $p^{l}$ is also a direction of descent for $\pazocal{H}^{l+1}$ at $\w^{l+1}_{\mu_1^{l+1}}$.

\paragraph{Level Convergence}
Given that the first-order consistency conditions \eqref{eq:cc_g} are imposed, all levels converge to the minimizer $\w^*$ of the original objective function $\pazocal{F}$, see \cite{Nash2000multigrid}.
Therefore, whenever $\nabla \pazocal{H}^{l+1}(\w^{l+1}) \rightarrow \nabla \pazocal{H}^{l+1}(\w^*) \rightarrow 0$, then also $\nabla \pazocal{H}^{l}(\w^{l+1}) \rightarrow 0$.

\begin{figure*} 
 \begin{tikzpicture}[]
    \begin{groupplot}[
        group style={
            group size = 4 by 1,
            horizontal sep=1.5pt,
            vertical sep=3cm,
          },
	legend pos=north east,
	ytick={1, 1e-3, 1e-6,  1e-9},
        width=0.285\textwidth, 
        height=0.235\textwidth, 
	grid=major, % Display a grid
	grid style={dashed,gray}, % Set the style
	xmode=normal,
	ymode=log,
	xlabel={\scriptsize  \# Grad / $n$},
	ymin=1e-10, 
	ymax=1,
	xmin=0,	
	tick label style={font=\scriptsize},
	label style={font=\scriptsize},
	legend style={font=\scriptsize}        
      ]

      \nextgroupplot[align=left, 
      title={},
      title={\scriptsize Mushrooms  ($\kappa=10^2$)},      
      ylabel={$\scriptsize \pazocal{F}(w) - \pazocal{F}(w^*)$},       
      xmax=65]
	\addplot[color = green2, very thick] table[x=grad_calls,y= loss_diff,col sep=comma] {mushrooms_MLVR_lev2_200.csv}; 
	
	\addplot[color = green3, very thick] table[x=grad_calls,y= loss_diff,col sep=comma] {mushrooms_MLVR_lev3_200.csv}; 
	
	\addplot[color = myred, very thick] table[x=grad_calls,y= g_norm,col sep=comma] {mushrooms_SARAH.csv}; 	
	
	\addplot[color = mypurple, very thick] table[x=grad_calls,y= g_norm,col sep=comma] {mushroomsSVRG.csv}; 
	
	\addplot[color = myyellow, very thick] table[x=grad_calls,y= g_norm, col sep=comma] {mushroomsSS_200.csv}; 

      \nextgroupplot[align=left, 
      title={},
      xmax=50,	
      title={\scriptsize Covtype ($\kappa=10^3$)},      
      yticklabels={}]
      
	\addplot[color = green2, very thick] table[x=grad_calls,y= loss_diff,col sep=comma] {covtype_MLVR_lev_2_5000.csv};   	
	\label{pgfplots:MLVR2}	
	
	\addplot[color = green3, very thick] table[x=grad_calls,y= loss_diff,col sep=comma] {covtype_MLVR_lev_3_5000.csv}; 
	\label{pgfplots:MLVR3}
	
	\addplot[color = myred, very thick] table[x=grad_calls,y= g_norm,col sep=comma] {covtypeSARAH.csv}; 
	\label{pgfplots:SARAH}
	
	\addplot[color = mypurple, very thick] table[x=grad_calls,y= g_norm,col sep=comma] {covtypeSVRG.csv}; 
	\label{pgfplots:SVRG}
	
	\addplot[color = myyellow, very thick] table[x=grad_calls,y= g_norm,col sep=comma] {covtypeSS_5000.csv}; 
	\label{pgfplots:SSN}

      \nextgroupplot[align=left, 
      title={},
     xmax=240,	      
      yticklabels={}, 
      title={\scriptsize Gisette ($\kappa=10^4$)}]
	
	\addplot[color = green2, very thick] table[x=grad_calls, y= loss_diff, col sep=comma] {gisette_MLVR_lev2_400.csv};    	    	
	
	\addplot[color = green3, very thick] table[x=grad_calls, y= loss_diff, col sep=comma] {gisette_MLVR_lev3_400.csv};     

	\addplot[color = myred, very thick] table[x=Grads_calls,y= Loss,col sep=comma] {gisette_SARAH.csv}; 
	
	\addplot[color = mypurple, very thick] table[x=Grads_calls,y= Loss,col sep=comma] {gisette_SVRG.csv}; 
	
	\addplot[color = myyellow, very thick] table[x=grad_calls, y= g_norm,col sep=comma] {gisetteSS_out__400.csv};

      \nextgroupplot[align=left, 
      title={},
      yticklabels={},       
      title={\scriptsize Australian ($\kappa=10^6$)} ,     
      	xmax= 50]

	\addplot[color = green2, very thick] table[x=grad_calls,y= loss_diff,col sep=comma] {australian_MLVR_lev_2_50.csv}; 
	
	\addplot[color = green3, very thick] table[x=grad_calls,y= loss_diff,col sep=comma] {australian_MLVR_lev_3_50.csv};
	
	\addplot[color = myred, very thick] table[x=Grads_calls,y= Loss,col sep=comma] {australian_SARAH.csv}; 
	
	\addplot[color = mypurple, very thick] table[x=Grads_calls,y= Loss,col sep=comma] {australian_SVRG.csv}; 
	
	\addplot[color = myyellow, very thick] table[x=grad_calls,y= g_norm,col sep=comma] {australianSS_out__100.csv};

    \end{groupplot}
    \matrix[ draw, matrix of nodes, anchor = west, node font=\scriptsize,
    column 1/.style={nodes={align=left,text width=0.9cm}},
    column 2/.style={nodes={align=center,text width=0.9cm}},
    ] at (current axis.east)
    {
   \scriptsize  SARAH & \ref{pgfplots:SARAH}  \\
   \scriptsize  SVRG & \ref{pgfplots:SVRG}  \\
   \scriptsize  SSN  & \ref{pgfplots:SSN}  \\
   \scriptsize  MLVR2   & \ref{pgfplots:MLVR2}   \\
    \scriptsize MLVR3   & \ref{pgfplots:MLVR3}   \\    
    };
  \end{tikzpicture}
  \caption{
  Training error, $\pazocal{F}(w)-\pazocal{F}(w^*)$, with respect to effective gradient evaluations for SVRG, SARAH, Sub-sampled Newton (SSN), two and three level variants of MLVR method (MLVR2, MLVR3).
  }
  \label{fig:test1}
\end{figure*}
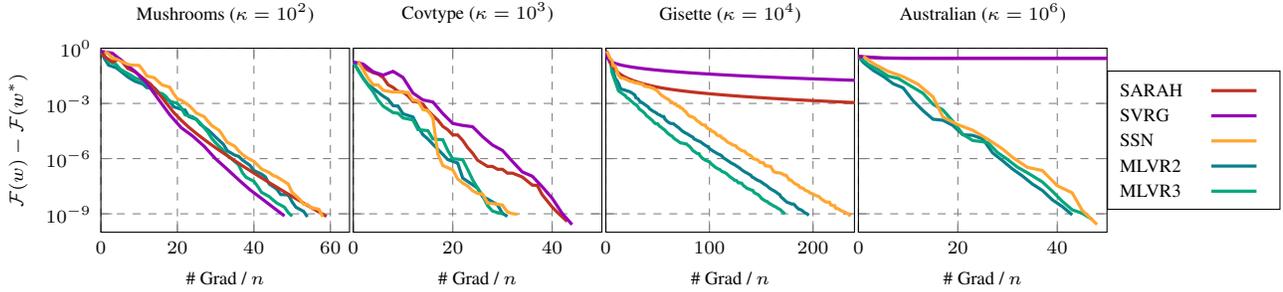

\paragraph{Variance Reduction}
Although, the gradient of the level dependent objective function $\pazocal{H}^l$, i.e $\nabla \pazocal{H}^l$,  is evaluated using  reduced dataset $\pazocal{D}^l$, its variance is lower compared to $\nabla \pazocal{F}^l$.
This is due to the fact that the coupling term $\delta g$, used to define $\pazocal{H}^l$ in \eqref{eq:coarse_objective} contains information about full gradient.

%%%%%%%%%%%%%%%%%%%%%%%%%%%%%%%%%%%%%%%%%
%%%%%%%%%%%%%%%%%%%%%%%%%%%%%%%%%%%%%%%%%
%%%%%%%%%%%%%%%%%%%%%%%%%%%%%%%%%%%%%%%%%
\subsection{Variants of MLVR}
\label{sec:connection}
MLVR method, Algorithm \ref{alg:mgopt}, is very generic as it can be configured in several ways.
Once a number of levels $L$ is chosen, the user can decide how to construct datasets $\{ \pazocal{D}^l \}_{l=1}^{l=L}$, which optimizer to use on every level and how many optimizers steps to perform.
This allows for the construction of many existing as well as many yet unexplored solution strategies. 
Here, we demonstrate that the particular variants of the two-level MLVR method already appear in machine learning literature.
In particular, we provide two examples, i.e. sub-sampled Newton and SVRG. 
 
\paragraph{Sub-sampled Newton}
Let us assume following setup, where MLVR is configured with $L=2, \mu^2_1=\mu^2_2=0, \mu^1=1$. 
The coarse level dataset is obtained as a subset of the full dataset, i.e. $\pazocal{D}^1 \subset \pazocal{D}$, and we employ Newton's method as an optimizer on the coarse level.
The V-cycle of MLVR method then produces the following update rule
\begin{align*}
\w_{i+1} = \w_{i} - \alpha  \bigg(\frac{1}{|\pazocal{D}^1|} \sum_{ j \in \pazocal{D}^1 } \nabla^2 f_j(\w_{i}) \bigg)^{-1}  \nabla \pazocal{F}(\w_{i}), 
\end{align*}
where $\alpha \in \R$. 
This update rule
 is known as a sub-sampled Newton (SSN) method, see for example \cite{berahas2020investigation, bollapragada2019exact}.

\paragraph{SVRG}
Let us assume MLVR setup, where $L=2, \mu^2_1=\mu^2_2=0, \mu^1=m$, where $m \in \N$. 
The coarse level dataset is identical to the full dataset, thus $\pazocal{D}^1 = \pazocal{D}$, and we employ the stochastic gradient (SGD) method as an optimizer on the coarse level.
The V-cycle of MLVR method then produces the following update rule
\begin{align}
\w_{i+1} = \w_{i} - \alpha  \big( \nabla f_{t_i}(\w_{i}) - \nabla f_{t_i}(\tilde{\w}) + \nabla \pazocal{F}(\tilde{\w}) \big), 
\label{eq:SVRG}
\end{align}
where $\alpha \in \R$, $t_i$ is chosen uniformly from $\{1, \dots, |\pazocal{D}| \}$, and $\tilde{\w}$ is so called snapshot vector.
In our MLVR method, $\tilde{\w}$ is obtained by returning to the fine level and subsequently constructing a coarse level objective function by means of \eqref{eq:coarse_objective}.
The update rule \eqref{eq:SVRG} was introduced in \cite{johnson2013accelerating} and it is well known under the name SVRG. 
Over the years, several extensions of SVRG  were proposed in the literature.
Some of them produce updates, which mimic closely standard techniques from multilevel literature. 
For example, authors of \cite{harikandeh2015stopwasting} propose to perform full gradient step every $m$ iterations. 
This can be understood as an equivalent to taking one pre-smoothing step ($\mu_1^2=1$ in Alg.~\ref{alg:mgopt}). 
Extension of SVRG using mini-batches was proposed in \cite{harikandeh2015stopwasting} and it can be interpreted as a special case of a three-level MLVR.% method.

\paragraph{Our MLVR}
As common in multilevel literature, we propose to setup MLVR as follows. 
On finer levels,  $l > 1$, where an evaluation of objective function and its derivatives is expensive, we employ only one iteration of gradient descent optimizer. 
In contrast, on the coarsest level, $l=1$, the function and its derivatives can be evaluated cheaply as only small subset of samples is considered. 
Hence, we perform one iteration of Newton's method, by using $10$ steps of Conjugate Gradient method for solving the linear system.

%%%%%%%%%%%%%%%%%%%%%%%%%%%%%%%%%%%%%%%%%
%%%%%%%%%%%%%%%%%%%%%%%%%%%%%%%%%%%%%%%%%
%%%%%%%%%%%%%%%%%%%%%%%%%%%%%%%%%%%%%%%%%
\section{Numerical Experiments}
\label{sec:num_results}
In this section, we illustrate the numerical performance of the proposed MLVR method on binary classification problems. 
Given a training set $\pazocal{D} = \{ (\x_i, \y_i) \}_{i=1}^n$,  we consider $\ell_2$-regularized logistic loss, defined as
\begin{align*}
\pazocal{F}(\w) := \frac{1}{n} \sum_{i=1}^{n} \log \big(1 + e^{- \y^i (\w^T \x^i)} \big) + \frac{\lambda}{2} \| w \|^2, 
\end{align*}
where $\lambda=\frac{1}{n}$ is a penalty parameter.
We consider four datasets, \emph{Australian}, \emph{Mushrooms}, \emph{Gisette}, \emph{Covtype}, from LIBSVM database \footnote{\url{https://www.csie.ntu.edu.tw/~cjlin/libsvmtools/datasets/binary.html}}, see Tab.~\ref{table:dataset_info} for the details regarding number of samples (n), the number of variables (d) and the condition number ($\kappa$). 
We compare the performance of MLVR method to the state-of-the-art variance reduction methods, SVRG and SARAH, and to the sub-sampled Newton's (SSN) method.
Fig.~\ref{fig:test1} illustrates the obtained results in terms of effective gradient evaluations, defined as the su=m of gradient evaluations and Hessian-vector products (required by the Conjugate Gradient method while solving linear systems). 
All methods were configured to the best performing variant by thorough hyper-parameter search, see supplementary material (Appendix \ref{app:details}) for the details. 
We consider zero initial guess and terminate the solution process, when the condition $\| \pazocal{F}(\w) - \pazocal{F}(\w^*)\|<10^{-9}$ is satisfied, where $\w^*$ denotes the minimizer.

\begin{table}
\centering
\footnotesize
\caption{\footnotesize Datasets}
\label{table:dataset_info}
\begin{tabular}{ |c|c|c|c| } 
 \hline
Dataset & $n$ &     $d$  & $\kappa$  \\ \hline  \hline
\emph{Mushroom} &  $6,499$ & $112$ & 	 $10^2$	\\ \hline
\emph{Cotypev} &   $406,708$  & $54$  &	 $10^3$ \\ \hline
\emph{Gisette} &   $6,000$  & $5,000$  &	$10^4$		\\ \hline
\emph{Australian} &   $621$ & $14$  &	$10^6$		\\ \hline
\end{tabular}
\end{table}

\textbf{Ill-conditioning}
Although the variance reduction methods (SVRG, SARAH) are very efficient for well-conditioned optimization problems (\emph{Mushroom}), their performance degrades for ill-conditioned optimization problems.
In contrast, methods that incorporate the second-order information, such as SSN and MLVR,  perform significantly better when the condition number of the Hessian ($\kappa$) increases.
For instance, they achieve more than $20$ times speedup compared to VR methods for  \emph{Gisette} and \emph{Australian} datasets.

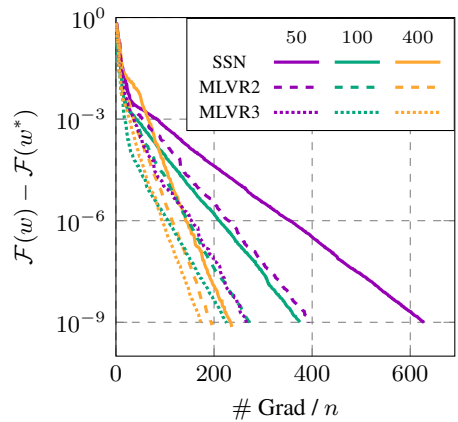
\begin{figure}[H]
\centering
 \begin{tikzpicture}
    \begin{axis}[
        width=0.355\textwidth, 
        height=0.355\textwidth,         
	ytick={1, 1e-3, 1e-6,  1e-9},
	grid=major, % Display a grid
	grid style={dashed,gray}, % Set the style
	xmode=normal,
	ymode=log,
	xlabel= {\footnotesize $\#$ Grad / $n$},
	ylabel= {\footnotesize $\pazocal{F}(w) - \pazocal{F}(w^*)$},
	ymin=1e-10, 
	ymax=1,
	xmin=0,	
	tick label style={font=\footnotesize},
      ]
      
	\addplot[color = mypurple, very thick] table[x=grad_calls,y= g_norm,col sep=comma] {gisetteSS_out__50.csv};      
	 \label{plot:SSN50}	
	\addplot[color = green3, very thick] table[x=grad_calls,y= g_norm,col sep=comma] {gisetteSS_out__100.csv};     
	 \label{plot:SSN100}  	
	\addplot[color = myyellow, very thick] table[x=grad_calls,y= g_norm,col sep=comma] {gisetteSS_out__400.csv};      
	\label{plot:SSN400}

	\addplot[color = mypurple, dashed, very thick] table[x=grad_calls,y= loss_diff,col sep=comma] {gisette_MLVR_lev2_50.csv};   	
	\label{plot:MLVR_l2_50}	
	\addplot[color = green3, dashed, very thick] table[x=grad_calls,y= loss_diff,col sep=comma] {gisette_MLVR_lev2_100.csv};  
	\label{plot:MLVR_l2_100}		
	\addplot[color = myyellow, dashed, very thick] table[x=grad_calls,y= loss_diff,col sep=comma] {gisette_MLVR_lev2_400.csv};    	    	
	\label{plot:MLVR_l2_400}

	\addplot[color = mypurple, densely dotted, very thick] table[x=grad_calls,y= loss_diff,col sep=comma] {gisette_MLVR_lev3_50.csv}; 
	\label{plot:MLVR_l3_50}			
	\addplot[color = green3, densely dotted, very thick] table[x=grad_calls,y= loss_diff,col sep=comma] {gisette_MLVR_lev3_100.csv};       
	\label{plot:MLVR_l3_100}				
	\addplot[color = myyellow, densely dotted, very thick] table[x=grad_calls,y= loss_diff,col sep=comma]{gisette_MLVR_lev3_400.csv};       
	\label{plot:MLVR_l3_400}

   \coordinate (legend) at (axis description cs:1.0, 0.665);
       
\end{axis}
{
\scriptsize
        \matrix [
            draw,
            matrix of nodes,
            anchor=south east,
            style={fill=white}
        ] at (legend) {
        
                    & $50$ & $100$ & $400$ \\
         \scriptsize  SSN & \ref{plot:SSN50}   & \ref{plot:SSN100} & \ref{plot:SSN400}  \\
         \scriptsize  MLVR2 & \ref{plot:MLVR_l2_50}   & \ref{plot:MLVR_l2_100} &  \ref{plot:MLVR_l2_400} \\
	\scriptsize MLVR3 & \ref{plot:MLVR_l3_50}   & \ref{plot:MLVR_l3_100}  & \ref{plot:MLVR_l3_400} \\
        };
}

\end{tikzpicture}
\caption{
Training error, $\pazocal{F}(w)-\pazocal{F}(w^*)$, with respect to effective gradient evaluations for  Sub-sampled Newton and MLVR methods for \emph{Gisette} dataset. 
}
\label{fig:gis_dataset}
\end{figure}

\newpage
\textbf{Multiple levels and sensitivity to hyper-parameters}
Introducing the hierarchy of multiple levels can be beneficial, in order to accelerate convergence, decrease computational cost, and reduce the sensitivity to the choice of hyper-parameters.
Fig.~\ref{fig:gis_dataset} demonstrates the performance of SSN and MLVR methods with a different number of samples used for evaluation of the sub-sampled Hessians. 
As we can see, the performance of the SSN method is more susceptible to the choice of hyper-parameters.
For instance, the SSN method with $50$ samples performs $2.6$ times worse than with $400$ samples.
In contrast, the performance of the MLVR3 method decreases only by a factor of $1.4$. 
Additionally, MLVR3 with $50$ samples achieves already performance comparable to a well-tuned SSN method (with 400 samples).

%%%%%%%%%%%%%%%%%%%%%%%%%%%%%%%%%%%%%%%%%
%%%%%%%%%%%%%%%%%%%%%%%%%%%%%%%%%%%%%%%%%
%%%%%%%%%%%%%%%%%%%%%%%%%%%%%%%%%%%%%%%%%
\section{Conclusion}
\label{sec:conclusion}
We proposed a novel training method, multilevel variance reduction (MLVR), which combines ideas from variance reduction and multilevel minimization techniques. 
We built a multilevel hierarchy by reducing the number of samples, which makes our method applicable to any type of machine learning problem.
Our preliminary numerical results suggest that the MLVR method outperforms standard variance reduction methods, especially when the underlying problem is ill-conditioned.
We also demonstrated that it is beneficial to explore multilevel hierarchy with more than two levels.

The presented work can be extended in many theoretical and empirical ways.
For example, we plan to investigate the numerical performance using larger datasets. 
We intend to explore non-uniform sub-sampling strategies in order to generate a hierarchy of datasets.
We also aim to combine coarsening in number of samples with coarsening in number of parameters, which could decrease the computational cost.

\bibliographystyle{icml2020}
\bibliography{paper.bib}

\begin{thebibliography}{28}
\providecommand{\natexlab}[1]{#1}
\providecommand{\url}[1]{\texttt{#1}}
\expandafter\ifx\csname urlstyle\endcsname\relax
  \providecommand{\doi}[1]{doi: #1}\else
  \providecommand{\doi}{doi: \begingroup \urlstyle{rm}\Url}\fi

\bibitem[Alexandrov \& Lewis(2001)Alexandrov and Lewis]{alexandrov2001overview}
Alexandrov, N.~M. and Lewis, R.~M.
\newblock An overview of first-order model management for engineering
  optimization.
\newblock \emph{Optimization and Engineering}, 2\penalty0 (4):\penalty0
  413--430, 2001.

\bibitem[Berahas et~al.(2020)Berahas, Bollapragada, and
  Nocedal]{berahas2020investigation}
Berahas, A.~S., Bollapragada, R., and Nocedal, J.
\newblock An investigation of newton-sketch and subsampled newton methods.
\newblock \emph{Optimization Methods and Software}, pp.\  1--20, 2020.

\bibitem[Bollapragada et~al.(2019)Bollapragada, Byrd, and
  Nocedal]{bollapragada2019exact}
Bollapragada, R., Byrd, R.~H., and Nocedal, J.
\newblock Exact and inexact subsampled newton methods for optimization.
\newblock \emph{IMA Journal of Numerical Analysis}, 39\penalty0 (2):\penalty0
  545--578, 2019.

\bibitem[Bottou et~al.(2018)Bottou, Curtis, and
  Nocedal]{bottou2018optimization}
Bottou, L., Curtis, F.~E., and Nocedal, J.
\newblock Optimization methods for large-scale machine learning.
\newblock \emph{Siam Review}, 60\penalty0 (2):\penalty0 223--311, 2018.

\bibitem[Brandt(1977)]{brandt1977multi}
Brandt, A.
\newblock Multi-level adaptive solutions to boundary-value problems.
\newblock \emph{Mathematics of computation}, 31\penalty0 (138):\penalty0
  333--390, 1977.
\newblock \doi{10.2307/2006422}.

\bibitem[Briggs et~al.(2000)Briggs, Henson, and McCormick]{Briggs2000multigrid}
Briggs, W.~L., Henson, V.~E., and McCormick, S.~F.
\newblock \emph{A multigrid tutorial}.
\newblock SIAM, second edition, 2000.
\newblock \doi{10.1137/1.9780898719505}.

\bibitem[Chang et~al.(2017)Chang, Meng, Haber, Tung, and
  Begert]{chang2017multi}
Chang, B., Meng, L., Haber, E., Tung, F., and Begert, D.
\newblock Multi-level residual networks from dynamical systems view.
\newblock \emph{arXiv preprint arXiv:1710.10348}, 2017.

\bibitem[Chen et~al.(2019)Chen, Hu, and Wise]{chenConvergenceAnalysisFast2019}
Chen, L., Hu, X., and Wise, S.~M.
\newblock Convergence {{Analysis}} of the {{Fast Subspace Descent Methods}} for
  {{Convex Optimization Problems}}.
\newblock \emph{arXiv:1810.04116 [cs, math]}, October 2019.
\newblock URL \url{http://arxiv.org/abs/1810.04116}.

\bibitem[Defazio et~al.(2014)Defazio, Bach, and
  Lacoste-Julien]{defazio2014saga}
Defazio, A., Bach, F., and Lacoste-Julien, S.
\newblock Saga: A fast incremental gradient method with support for
  non-strongly convex composite objectives.
\newblock In \emph{Advances in neural information processing systems}, pp.\
  1646--1654, 2014.

\bibitem[Gaedke-Merzh{\"a}user et~al.(2020)Gaedke-Merzh{\"a}user,
  Kopani{\v{c}}{\'a}kov{\'a}, and Krause]{gaedke2020multilevel}
Gaedke-Merzh{\"a}user, L., Kopani{\v{c}}{\'a}kov{\'a}, A., and Krause, R.
\newblock Multilevel minimization for deep residual networks.
\newblock \emph{arXiv preprint arXiv:2004.06196}, 2020.

\bibitem[Goodfellow et~al.(2016)Goodfellow, Bengio, and
  Courville]{goodfellow2016deep}
Goodfellow, I., Bengio, Y., and Courville, A.
\newblock \emph{Deep learning}.
\newblock MIT press, 2016.

\bibitem[Gratton et~al.(2008)Gratton, Sartenaer, and
  Toint]{Gratton2008recursive}
Gratton, S., Sartenaer, A., and Toint, P.~L.
\newblock {Recursive Trust-Region Methods for Multiscale Nonlinear
  Optimization}.
\newblock \emph{SIAM Journal on Optimization}, 19\penalty0 (1):\penalty0
  414--444, 2008.
\newblock \doi{10.1137/050623012}.

\bibitem[Gro{\ss} \& Krause(2009)Gro{\ss} and Krause]{Gross2009}
Gro{\ss}, C. and Krause, R.
\newblock {On the Convergence of Recursive Trust-Region Methods for Multiscale
  Nonlinear Optimization and Applications to Nonlinear Mechanics}.
\newblock \emph{SIAM Journal on Numerical Analysis}, 47\penalty0 (4):\penalty0
  3044--3069, 2009.
\newblock \doi{10.1137/08071819X}.

\bibitem[Hackbusch(1985)]{hackbusch2013multi}
Hackbusch, W.
\newblock \emph{Multi-grid methods and applications}, volume~4.
\newblock Springer-Verlag Berlin Heidelberg, 1985.
\newblock \doi{10.1007/978-3-662-02427-0}.

\bibitem[Harikandeh et~al.(2015)Harikandeh, Ahmed, Virani, Schmidt,
  Kone{\v{c}}n{\`y}, and Sallinen]{harikandeh2015stopwasting}
Harikandeh, R.~B., Ahmed, M.~O., Virani, A., Schmidt, M., Kone{\v{c}}n{\`y},
  J., and Sallinen, S.
\newblock Stopwasting my gradients: Practical svrg.
\newblock In \emph{Advances in Neural Information Processing Systems}, pp.\
  2251--2259, 2015.

\bibitem[Hovhannisyan et~al.(2016)Hovhannisyan, Parpas, and
  Zafeiriou]{hovhannisyan2016magma}
Hovhannisyan, V., Parpas, P., and Zafeiriou, S.
\newblock Magma: Multilevel accelerated gradient mirror descent algorithm for
  large-scale convex composite minimization.
\newblock \emph{SIAM Journal on Imaging Sciences}, 9\penalty0 (4):\penalty0
  1829--1857, 2016.

\bibitem[Johnson \& Zhang(2013)Johnson and Zhang]{johnson2013accelerating}
Johnson, R. and Zhang, T.
\newblock Accelerating stochastic gradient descent using predictive variance
  reduction.
\newblock In \emph{Advances in neural information processing systems}, pp.\
  315--323, 2013.

\bibitem[Kone{\v{c}}n{\`y} \& Richt{\'a}rik(2017)Kone{\v{c}}n{\`y} and
  Richt{\'a}rik]{konevcny2017semi}
Kone{\v{c}}n{\`y}, J. and Richt{\'a}rik, P.
\newblock Semi-stochastic gradient descent methods.
\newblock \emph{Frontiers in Applied Mathematics and Statistics}, 3:\penalty0
  9, 2017.

\bibitem[Kopani\v{c}\'akov\'a \& Krause(2019)Kopani\v{c}\'akov\'a and
  Krause]{kopanicakova_2019c}
Kopani\v{c}\'akov\'a, A. and Krause, R.
\newblock A recursive multilevel trust region method with application to fully
  monolithic phase-field models of brittle fracture.
\newblock \emph{Computer Methods in Applied Mechanics and Engineering}, pp.\
  112720, 2019.
\newblock ISSN 0045-7825.
\newblock \doi{https://doi.org/10.1016/j.cma.2019.112720}.

\bibitem[Kopani\v{c}\'akov\'a et~al.(2019)Kopani\v{c}\'akov\'a, Krause, and
  Tamstorf]{kopanicakova2019b}
Kopani\v{c}\'akov\'a, A., Krause, R., and Tamstorf, R.
\newblock Subdivision-based nonlinear multiscale cloth simulation.
\newblock \emph{SIAM Journal on Scientific Computing}, 41\penalty0
  (5):\penalty0 S433--S461, 2019.
\newblock \doi{10.1137/18M1194870}.
\newblock URL \url{https://doi.org/10.1137/18M1194870}.

\bibitem[Kornhuber \& Krause(2001)Kornhuber and
  Krause]{kornhuber_adaptive_2001}
Kornhuber, R. and Krause, R.
\newblock Adaptive {{Multigrid Methods}} for {{Signorini}}'s {{Problem}} in
  {{Linear Elasticity}}.
\newblock \emph{Computing and Visualization in Science}, 4\penalty0
  (1):\penalty0 9--20, 2001.

\bibitem[Mairal(2013)]{mairal2013optimization}
Mairal, J.
\newblock Optimization with first-order surrogate functions.
\newblock In \emph{International Conference on Machine Learning}, pp.\
  783--791, 2013.

\bibitem[Nash(2000)]{Nash2000multigrid}
Nash, S.~G.
\newblock {A multigrid approach to discretized optimization problems}.
\newblock \emph{Optimization Methods and Software}, 14\penalty0 (1-2):\penalty0
  99--116, 2000.
\newblock \doi{10.1080/10556780008805795}.

\bibitem[Nguyen et~al.(2017)Nguyen, Liu, Scheinberg, and
  Tak{\'a}{\v{c}}]{nguyen2017sarah}
Nguyen, L.~M., Liu, J., Scheinberg, K., and Tak{\'a}{\v{c}}, M.
\newblock Sarah: A novel method for machine learning problems using stochastic
  recursive gradient.
\newblock In \emph{Proceedings of the 34th International Conference on Machine
  Learning-Volume 70}, pp.\  2613--2621. JMLR. org, 2017.

\bibitem[Robbins \& Monro(1951)Robbins and Monro]{robbins1951stochastic}
Robbins, H. and Monro, S.
\newblock A stochastic approximation method.
\newblock \emph{The annals of mathematical statistics}, pp.\  400--407, 1951.

\bibitem[Schmidt et~al.(2017)Schmidt, Le~Roux, and Bach]{schmidt2017minimizing}
Schmidt, M., Le~Roux, N., and Bach, F.
\newblock Minimizing finite sums with the stochastic average gradient.
\newblock \emph{Mathematical Programming}, 162\penalty0 (1-2):\penalty0
  83--112, 2017.

\bibitem[Tai \& Xu(2002)Tai and Xu]{tai2002global}
Tai, X.-C. and Xu, J.
\newblock Global and uniform convergence of subspace correction methods for
  some convex optimization problems.
\newblock \emph{Mathematics of Computation}, 71\penalty0 (237):\penalty0
  105--124, 2002.

\bibitem[Yavneh \& Dardyk(2006)Yavneh and Dardyk]{yavneh2006multilevel}
Yavneh, I. and Dardyk, G.
\newblock A multilevel nonlinear method.
\newblock \emph{SIAM journal on scientific computing}, 28\penalty0
  (1):\penalty0 24--46, 2006.

\end{thebibliography}

\clearpage
\appendix
\section{Solution Strategies Setup}
\label{app:details}
In this section, we report details regarding the hyper-parameters setup, which was used to produce numerical results in Section \ref{sec:num_results}. 
Table \ref{tab:param_VR} specifies the step size $\alpha \in \R$ used by  SVRG, SARAH and SSN methods.
Since SVRG and SARAH operate in outer-inner mode, we also specify the number of inner iterations, denoted by $m$. 
As standard in the literature, we show the number of inner iterations by means of the number of samples of the full dataset $n$. 

Two and three level variants of MLVR method employ hierarchy of datasets $\{\pazocal{D}\}_{l=1}^{l=L}$, where $L=\{2, 3\}$.
The number of samples associated with each dataset is depicted in Table \ref{tab:param_ML_hierarchy}.
As we can see, the finest level, $l=L$, is always associated with the full dataset, thus $|\pazocal{D}^L| = n$. 
The coarsest level, $l=1$, is chosen, such that it contains the same amount of samples, as used for the evaluation of approximate Hessian by the SSN method. 
The number of samples associated with intermediate levels was obtained by doubling the number of samples from the subsequent coarser level. 
We also remark, that in our experiments, the SSN method employs a full gradient.
Thus, a sub-sampling strategy is used only for the evaluation of the approximate Hessian.

MLVR employs optimizer of the user choice on each level of the multilevel hierarchy.
In this work, we employ gradient descent (GD) optimizer on all levels, except on the coarsest, where one step of Newton's method is performed, thus $\mu^1=1$ and $\mu^l_1=1, \mu_2^l=0$, for all $l \in \{2, \dots, L\}$. 
Both methods, SSN and MLVR, employ a simple backtracking line-search method, in order to determine step size.

SSN method as well as the MLVR method (on the coarsest level) requires the solution of a linear system. 
In this work, we solve arising linear systems only approximately, by employing $10$ iterations of the Conjugate Gradient (CG) method. 
As the CG method requires only matrix-vector products instead of actually matrix, we do not assemble sub-sampled Hessian explicitly.
We rather perform Hessian-vector products directly.
Given our objective function, regularized logistic regression, the cost of performing the Hessian-vector product is equivalent to the cost of computing the gradient. 
We take into account this fact while reporting numerical results in terms of the number of effective gradient evaluations, see Section \ref{sec:num_results}.

\begin{table}[H]
\centering
\footnotesize
\caption{
\footnotesize
The number of samples used to build a multilevel hierarchy of datasets for MLVR2 and MLVR3 methods. 
The number of samples used to construct Hessian approximation within the SSN method. 
}
\label{tab:param_ML_hierarchy}
\begin{tabular}{ |c|c|c|c|c|c| }
 \hline
%Method & MLVR2 \big($|\pazocal{D}^1|$; $|\pazocal{D}^2|$ \big) & MLVR3 \big($|\pazocal{D}^1|$; $|\pazocal{D}^2|$; $|\pazocal{D}^3|$ \big) & SNN \\ \hline  \hline
Method & MLVR2 & MLVR3 & SNN \\ \hline  \hline
\emph{Australian} 	& (100; n) 		& (100; 200; n) & 100 \\ \hline
\emph{Gisette} 		& (400; n) 		& (400; 800; n) & 400 \\ \hline
\emph{Mushrooms} 	& (200; n) 		& (200; 400; n)  & 200  \\ \hline
\emph{Covtype}  	& (5,000; n)  	& (5,000; 10,000; n)  & 5,000 \\ \hline
\end{tabular}
\end{table}

\begin{table}[H]
\footnotesize
\centering
\caption{
\footnotesize
Set of parameters used during numerical experiments for SVRG and SARAH.
}
\label{tab:param_VR}
\begin{tabular}{ |c|c|c|c|c|c|c|c|c|c|c|c|c| } 
 \hline
Dataset & \multicolumn{2}{c|}{{\emph{Australian}}} & \multicolumn{2}{c|}{{\emph{Gisette}}}  \\ \hline
Method & $\alpha$  & $m$ & $\alpha$ & $m$    \\ \hline  \hline
SVRG &   $10^{-7}$  &  $5n$  &   $10^{-5}$ &  $n/2$ 	\\ \hline
SARAH &  $10^{-8}$  & $n/2$ &  $10^{-4}$   & $n/2$		\\ \hline \hline
%\end{tabular}
%\begin{tabular}{ |c|c|c|c|c|c|c|c|c|c|c|c|c| } 
% \hline
Dataset & \multicolumn{2}{c|}{{\emph{Mushrooms}}} & \multicolumn{2}{c|}{{\emph{Covtype}}} \\ \hline
Method & $\alpha$ &  $m$ & $\alpha$ &  $m$      \\ \hline  \hline
SVRG &   $0.5$  & $n$  &   $1$  & $0.5n$ 					\\ \hline
SARAH &  $0.1$ &  $0.5n$	&  $1$  & $2n$			\\ \hline
\end{tabular}
\end{table}

\end{document}